\documentclass{esannV2}
\usepackage{url}
\usepackage{multirow}
\usepackage{graphicx}
\usepackage{latexsym}
\usepackage{booktabs}
\usepackage{subcaption}  
\usepackage{xcolor}
\usepackage{flushend}
\usepackage{amsmath}

\usepackage[a4paper,
    left=4.4cm,
    right=4.4cm,
    top=5.2cm,
    bottom=5.2cm
]{geometry}

\begin{document}

\title{Safer Prompts: Reducing Risks from Memorization in Visual Generative AI}

\author{Lena Reißinger$^1$, Yuanyuan Li$^2$, Anna-Carolina Haensch$^1$, Neeraj Sarna $^2$
%
\vspace{.3cm}\\
%
1- Ludwig Maximilian University of Munich \\
Geschwister-Scholl-Platz 1, Munich, Germany
%
\vspace{.1cm}\\
2- Munich RE \\
Koeniginstr. 107, Munich, Germany\\
}

\maketitle
\begin{abstract}
Visual Generative AI models have demonstrated remarkable capability in generating high-quality images from user inputs like text prompts. However, because these models have billions of parameters, they risk memorizing certain parts of the training data and reproducing the memorized content. Memorization often raises concerns about safety of such models---usually involving intellectual property (IP) infringement risk---and deters their large scale adoption. In this paper, we evaluate the effectiveness of prompt engineering techniques in reducing memorization risk in image generation. Our findings demonstrate the effectiveness of prompt engineering in reducing the similarity between generated images and the training data of diffusion models, while maintaining relevance and aestheticity of the generated output. 
\end{abstract}

\section{Introduction}
\label{sec:intro}
As generative AI (GenAI) becomes increasingly prevalent in real-world applications, concerns about its potential risks continue to grow. We focus on the risks associated with so-called memorization where the model \textit{memorizes} the training data and reproduces a similar copy \cite{carlini_extracting_2023}. 
Since large scale models are trained on datasets that usually contain copyrighted material, memorization of training data leads to concerns around Intellectual Property (IP) violation, which some AI developers have already experienced \cite{andersen2023copyright}.

Risks associated with memorization not only deter a wide scale adoption of GenAI models but also hinder model development where an AI developer might sacrifice output quality at the expense of using limited training data. To promote a wider adoption and a safer development of GenAI, risk mitigation is crucial. We briefly review the available risk mitigation strategies.

Our focus is on post-deployment strategies that work solely with the model output (i.e., they do not require an access to the model weights or the model training pipeline) and are usually cheaper. One possibility is to add a "system message" to the user-prompt that aims to reduce the IP-infringement risks \cite{Microsoft_Copyright_Commitment}; we recall that similar prompt engineering techniques have been extensively used to enhance GenAI model performance on diverse tasks \cite{qiao2022reasoning,tonmoy2024comprehensive}. Another approach is to  use VLLMs to detect prompts that might generate
copyrighted images. In case such a prompt is detected, the diffusion process is guided away from copyrighted outputs by conditioning on trigger words \cite{sonyAI_evaluation}. Furthermore, prompt re-writing is also effective when combined with negative prompting \cite{fantastic_beasts}. 

\textbf{Current Contributions:}
We focus on prompt engineering and evaluate its effectiveness for memorization risk reduction. 
We hypothesis that via a carefully engineered prompt, memorization risks can be reduced. To the best of our knowledge, for vision generation models, the use of prompt engineering for memorization risk reduction is largely unexplored. Following is a summary of our contributions: i) we evaluate the generated output on three criteria that capture memorization risk and image quality; ii) under the aforementioned metrics, we evaluate four different prompt engineering strategies summarized in section \ref{sec:prompt strategies}; and iii) while deriving insights from our experiments, we conclude with practical recommendations for safer usage of visual generation models.  


\section{Prompting Strategies} \label{sec:prompt strategies}
We consider the following four prompting strategies.

    (i) \textit{Baseline/No prompt engineering:} here we directly use the captions of the training images to generate the outputs. We consider this to be a baseline strategy. The prompt for this strategy reads: \textit{Generate an image of $\{caption\}$}.

(ii) \textit{Task instruction prompting:} involves adding in the prompt a very detailed description of the task the model should perform. For mitigating memorization risk, this includes steering the model towards creating novel elements to produce unique output, as well as avoiding the reproduction of recognizable content. The prompt for this strategy reads: \textit{Create a visually distinctive, highly creative, and non-copyright-infringing depiction of $\{caption\}$. Focus on originality and incorporate entirely novel visual elements. Avoid using recognizable characters, logos, or copyrighted designs. Ensure the image is imaginative and unique. }

(iii) \textit{Negation prompting:} This includes the concept of negation (no, not, nor) within the (baseline) hard prompt. 
 The effect of this strategy on stable diffusion has already been explored \cite{li_get_2024}. We study its effectiveness in reducing memorization risk. The prompt reads: \textit{Generate an imaginative and original image of $\{caption\}$. The image must not include realistic replication, no known art styles, no recognizable characters, and no copyrighted material.}

(iv) \textit{Chain-of-thought prompting:} This enables the model with self-check mechanisms where a model evaluates its reasoning. This could potentially improve model's ability to generate unique and non-infringing images as outputs. The prompt reads: \textit{1. Generate a creative and unique image of $\{caption\}$, focusing on originality and imaginative composition. 2. Incorporate completely novel elements into the image that are distinct from the training data and are unlikely to resemble any existing images. 3. Ensure every element in the image is visually distinct, creative, and does not replicate known styles, characters, or objects present in existing datasets. 4. Verify the final output aligns with the given caption while maintaining a high degree of creativity and uniqueness.}

 The specific wordings of the aforementioned prompts were refined through a trial-and-error process during our initial tests. While this method may not be entirely systematic, informal trial-and-error approaches, as described by \cite{liu_design_2022}, have so far been the primary way prompts for text-to-image models have been developed. 


\section{Experimental Results}
\textbf{Evaluation Criteria:}\label{sec:IP detection}
We evaluate the generated output on three criteria: a) similarity to training images; b) relevance to the input prompt; and c) aestheticity.
Our goal is to reduce memorization while maintaining relevance to the user input and aestheticity of the generated output. 

With $E(\cdot)$, we represent a CLIP \cite{radford_learning_2021} encoder that could encode both a prompt $P$ and an image $X$ in the same space. To measure the similarity between two images $X_1$ and $X_2$, we use the cosine-similarity between the encodings $E(X_1)$ and $E(X_2)$. We represent this cosine-similarity by $sim(X_1,X_2)$. Two images $X_1$ and $X_2$ are \textit{similar} to each other when $sim(X_1,X_2) \geq \tau$, following prior work \cite{carlini_extracting_2023}, we use $\tau = 0.85$. Note that CLIP captures the content of an image and not necessarily its style. We do not focus on style because it might not be considered copyrighted thereby, resulting in low risk from memorization \cite{murray_generative_2023}.
The consine similarity between $E(X)$ and $E(P)$ measures the relevance of a prompt $P$ to an image $X$; we represent it using $rel(X,P)$. To measure aestheticity, we input the image $X$ into LAION-Aesthetics V2 predictor \cite{schuhmann_aesthetic_predictor}. Denoting the predictor using $\mathcal{A}$, the aestheticity score reads $aes(X) := \mathcal{A}(X).$

\textbf{Model and dataset:} Similar to \cite{carlini_extracting_2023}, we use Stability AI's Stable Diffusion 2 \cite{rombach_stabilityaistable-diffusion-2_2022} as an example. As training dataset, we consider the set LAION-Aesthetics 12M \cite {noauthor_dclurelaion-aesthetics-12m-umap_nodate}, which is the subset of the entire training set LAION-2B-en with aesthetics scores of 6 or higher \cite{rombach_stabilityaistable-diffusion-2_2022}. We refer to this caption-image set as $\mathcal{D}_t$.

\textbf{Prompt sampling strategy:} From $\mathcal{D}_t$, we extract captions, which when used as prompts, generate highly memorized images. We study the effect of our prompting strategy on only these \textit{high} risk captions. To extract these captions, we randomly sample 5000 caption-image pair from $\mathcal{D}_t$. Out of these captions, we choose the ones that, when fed into the model, generate images that are \textit{similar} to the images in $\mathcal{D}_t$. This results in 67 captions. For each prompt, we consider 75 different initializations. The choice of 75 generations strikes balance between computational expense and representativeness \cite{carlini_extracting_2023,liu_design_2022}. Repeating this process for $67$ prompts, we get 20,100 generated images (67$\times$75$\times$4). 



\begin{figure}[htbp!] 
    \centering
    \begin{subfigure}{0.45\textwidth}
            \centering
    \includegraphics[width=0.9\linewidth]{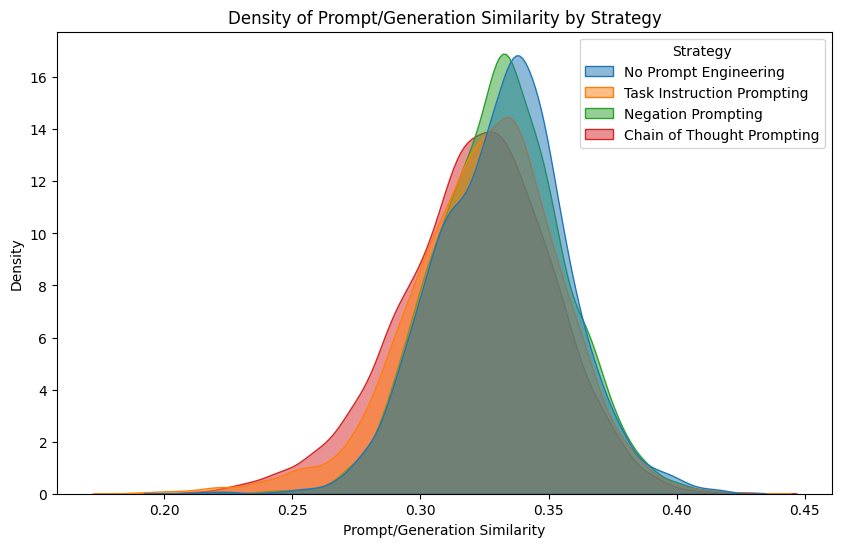}
    \caption{Relevance score distribution.}
    \label{fig:clip}
    \end{subfigure}
    \hfill
    \begin{subfigure}{0.45\textwidth}
    \centering
    \includegraphics[width=0.9\linewidth]{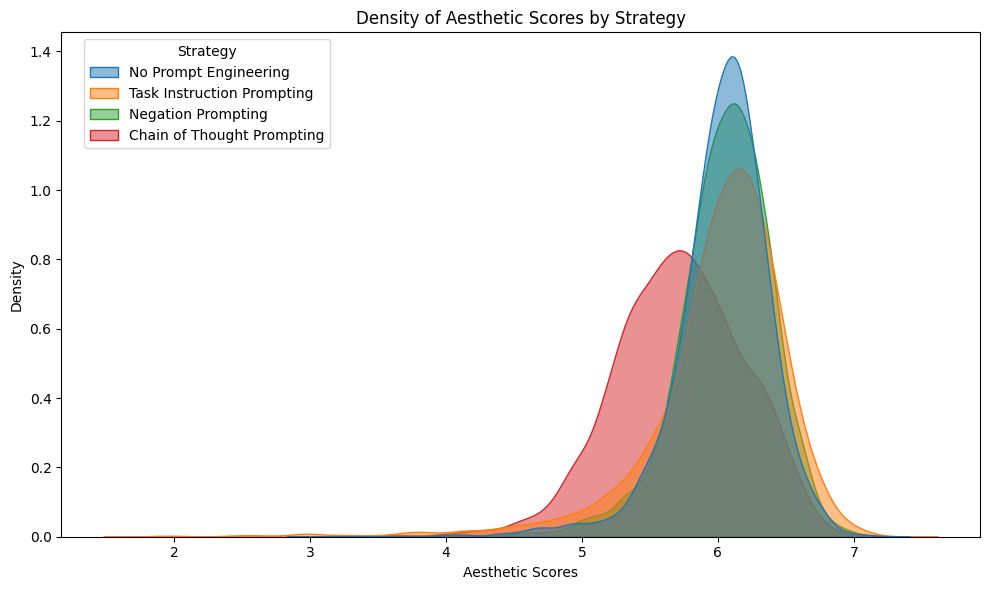}
    \caption{Aesthetic score distribution.}
    \label{fig: aesthetic}
    \end{subfigure}
    \caption{Quality assessment of generated images across prompting strategies.}
\end{figure}

\textbf{Memorization Reduction:}
We study the likelihood of generating images that are similar to those contained in our training set $\mathcal{D}_t$. Table \ref{tab:combined_similarity} highlights that prompt engineering is particularly effective in reducing memorization risk. It reduces the fraction of generated images that are similar to the training data. Without any prompt engineering, a total of 2,082 images (41.4$\%$ of 20,100 generated images) are similar to the images in $\mathcal {D}_t$. Then comes Negation prompting that reduces this number to 1751, which is further lowered to 1026 by Task Instruction prompting. The most effective strategy, chain-of-thought prompting, reduced this number to only 484 images. Next, we consider the mean similarity scores (taken over 75 samples) per prompt. Without prompt engineering, 21 prompts produced generations with mean similarity scores above 0.85. Negation prompting reduced this to 16, task-instruction prompting lowered it further to 7, and the most effective method—chain-of-thought prompting—brought it down to just one. In other words, on average, only one of the 67 tested prompts generated an image similar to those in the training set, $\mathcal{D}_t$.

\begin{table}[htbp] 
    \centering
    \caption{Frequency of generations (prompts) that are highly similar to training data}
    \label{tab:combined_similarity}

    \begin{tabular}{lccc}
        \toprule
    \textbf{Prompt Engineering Strategy} & {\textbf{Count}} & {\textbf{Frequency}} \\
    & Gen. (Prmpt.) & Gen. (Prmpt.) \\
    \midrule
        No Prompt Engineering        & 2082 (21)  & 41.43 (31.34) \\
        Task Instruction Prompting  & 1026 (7)  & 20.42 (10.45) \\
        Negation Prompting  & 1751 (16)  & 34.85 (23.88) \\
        Chain of Thought Prompting  & 484 (1)   & 9.63 (1.49)  \\
        \bottomrule
    \end{tabular}


\end{table}


\textbf{Relevance to input prompts:} To evaluate the generated image wrt. the input prompt, we compute the relevance score $rel(X,P)$, with $X$ being the generated image and $P$ being the base prompt without any prompt-engineering. Figure \ref{fig:clip} presents the results. Chain-of-thought prompting shows a slightly wider spread, which suggests more variability in how closely the generated images align with their original prompts. Negation prompting and no prompt engineering show slightly higher peak densities which implies more consistent alignment with the original captions. These findings suggest that prompt engineering influence generation outcomes with limited impact on prompt-image relevance.

\textbf{Aesthetic quality: }Figure \ref{fig: aesthetic} presents the aesthetic scores. Images generated without prompt engineering and negation prompting have the highest aesthetic scores with peaks around 6.2 - 6.3. Scores above 5 are generally considered favorable from an aesthetic perspective \cite{LAION_Aesthetics_V1}. Task instruction prompting  shows a broader distribution, suggesting greater variability in aesthetic quality. In contrast, chain-of-thought prompting yields noticeably lower aesthetic scores, suggesting that while this approach may reduce memorization risks, it does so at the cost of reduced aesthetic appeal. Nevertheless, most scores still exceed 5, indicating that the overall image quality remains acceptable.

\textbf{Correlation between memorization and image quality:}
To quantify how memorization relates to image quality, we compute the Pearson coefficient $r$ between the maximum similarity score (across different initializations) and the two attributes: aesthetic score $aes(X)$ and relevance score $rel(X,P)$ for each prompting strategy. Chain-of-thought prompting shows the strongest positive correlation with both aesthetics ($r$ = 0.49) and relevance ($r$ = 0.33), indicating that higher memorization risk often yields more pleasing and prompt-aligned images. Task instruction prompting has weaker correlations ($r$ = 0.25 for relevance and $r$ = 0.13 for aesthetics), while other strategies show negligible relationships (less than 0.15). Overall, reducing memorization risk — especially for Chain-of-thought — may slightly compromise image quality and relevance.

\textbf{Practical recommendations:}
We envision three different categories of applications: a) high risk; b) medium risk; and c) low risk. The higher the probability of memorization leading to financial losses, the higher is the risk for an application. For high-risk scenarios we recommend the Chain-of-Through prompting strategy with the highest memorization risk reduction. For medium-risk applications, Task instruction prompting could be preferable because it balances memorization risk with the quality of the generated image. For low-risk applications, Negation prompting is recommended,which provides the most relevant and aesthetically pleasing outputs while offering moderate memorization risk reduction.

\section{Conclusions}
We evaluate the effectiveness of prompting strategies in reducing the memorization risk of visual GenAI. Overall, we find that prompt engineering can reduce memorization risks in visual GenAI models, but its effectiveness varies depending on the chosen technique. Chain-of-thought prompting proved to be the most effective in memorization risk mitigation. Negation prompting was the least effective strategy, while task instruction prompting yielded promising results while nicely balancing memorization reduction with superior image quality.

\end{document}